\pdfoutput=1
\documentclass[11pt,a4paper]{article}
\usepackage{arxiv}
\usepackage[utf8]{inputenc} 
\usepackage[T1]{fontenc}    
\usepackage{amsfonts}       
\usepackage{nicefrac}       
\usepackage{lipsum}		
\usepackage{natbib}
\usepackage{doi}

\usepackage{times}
\usepackage{latexsym}
\usepackage{xspace}
\usepackage{multirow}
\usepackage{url}
\usepackage{booktabs}
\usepackage{tikz,tikz-qtree}
\usepackage{pgfplots}
\usepackage{amssymb}
\usepackage{xfrac}
\usepackage{graphicx}
\usepackage{tablefootnote}
\usepackage{amsmath}
\usepackage{enumitem}
\usepackage{todonotes}
\usepackage{subfigure}
\pgfplotsset{compat=1.14}

\usepackage{microtype}

\relax

\definecolor{g-red}{HTML}{DB4437}
\definecolor{g-blue}{HTML}{4285F4}
\definecolor{g-green}{HTML}{0F9D58}
\definecolor{g-yellow}{HTML}{F4B400}
\definecolor{g-orange}{HTML}{FF9800}
\definecolor{g-grey}{HTML}{9E9E9E}
\setenumerate[1]{itemsep=0pt,partopsep=0pt,parsep=\parskip,topsep=5pt}
\setitemize[1]{itemsep=0pt,partopsep=0pt,parsep=\parskip,topsep=5pt}
\setdescription{itemsep=0pt,partopsep=0pt,parsep=\parskip,topsep=5pt}

\renewcommand{\v}[1]{\boldsymbol{#1}}
\renewcommand{\t}[1]{\text{#1}}

\DeclareMathOperator*{\softmax}{softmax}

\DeclareMathOperator{\linear}{Linear}
\DeclareMathOperator{\layernorm}{LayerNorm}

\DeclareMathOperator{\multihead}{MultiHead}
\DeclareMathOperator{\mlp}{MLP}
\DeclareMathOperator{\attblok}{AttentionBlock}
\DeclareMathOperator{\mask}{MASK}
\DeclareMathOperator{\mentionatt}{MentionAttention}
\DeclareMathOperator{\neighboratt}{NeighborAttention}
\DeclareMathOperator{\learnscore}{LearnScore}

\title{\textit{BoningKnife}: Joint Entity Mention Detection and Typing for Nested NER via prior Boundary Knowledge}
\renewcommand{\shorttitle}{\textit{BoningKnife}: Joint Entity Mention Detection and Typing for Nested NER via prior ...}

\author{
Huiqiang Jiang\textsuperscript{$^\clubsuit$},
Guoxin Wang\textsuperscript{$^\diamondsuit$},
Weile Chen\textsuperscript{$^\heartsuit$},
Chengxi Zhang\textsuperscript{$^\clubsuit$},
B{\"o}rje F. Karlsson\textsuperscript{$^\blacklozenge$}\\

\textsuperscript{$^\clubsuit$}Peking University,
\textsuperscript{$^\blacklozenge$}Microsoft Research\\
\textsuperscript{$^\diamondsuit$}Microsoft Azure AI, \textsuperscript{$^\heartsuit$}Harbin Institute of Technology\\
\{jhq, chengxi\_zhang\}@pku.edu.cn, \{guow, borjekar\}@microsoft.com, chen.weile7@gmail.com\\
}

\date{}

\begin{document}
\maketitle
\begin{abstract}
    While named entity recognition (NER) is a key task in natural language processing, most approaches only target flat entities, ignoring nested structures which are common in many scenarios.
Most existing nested NER methods traverse all sub-sequences which is both expensive and inefficient, and also don't well consider boundary knowledge which is significant for nested entities.
In this paper, we propose a joint entity mention detection and typing model via prior boundary knowledge (\textit{BoningKnife}) to better handle nested NER extraction and recognition tasks.
\textit{BoningKnife} consists of two modules, MentionTagger and TypeClassifier.
MentionTagger better leverages boundary knowledge beyond just entity start/end to improve the handling of nesting levels and longer spans, while generating high quality mention candidates. TypeClassifier utilizes a two-level attention mechanism to decouple different nested level representations and better distinguish entity types. We jointly train both modules sharing a common representation and a new dual-info attention layer, which leads to improved representation focus on entity-related information.
Experiments over different datasets show that our approach outperforms previous state of the art methods and achieves 86.41, 85.46, and 94.2 F1 scores on ACE2004, ACE2005, and NNE, respectively.\footnote{Work performed at Microsoft Research Asia between 2019/2020.}

\end{abstract}
\section{Introduction}
\label{sec:Introduction}
Named Entity Recognition (NER) is a fundamental tasks in natural language processing (NLP), which aims to extract and recognize named entities, like person names, organizations, geopolitical entities, etc., in unstructured text.
However, in addition to flat entity mentions, nested or overlapping entities are commonplace in natural language.
Such nested entities bring richer entity knowledge and semantics and can be critical to facilitate various downstream NLP tasks and real-world applications.
As an example of their frequency, nested entities account for 35.19\%, 30.80\%, and 66.14\% of mentions in standard datasets like ACE2004 \cite{doddington2004automatic}, ACE2005 \cite{walker2006ace}, and NNE \cite{dblp2019nne}, respectively.

\begin{figure}[t!]
    \begin{center}
    \includegraphics[width=0.7\textwidth]{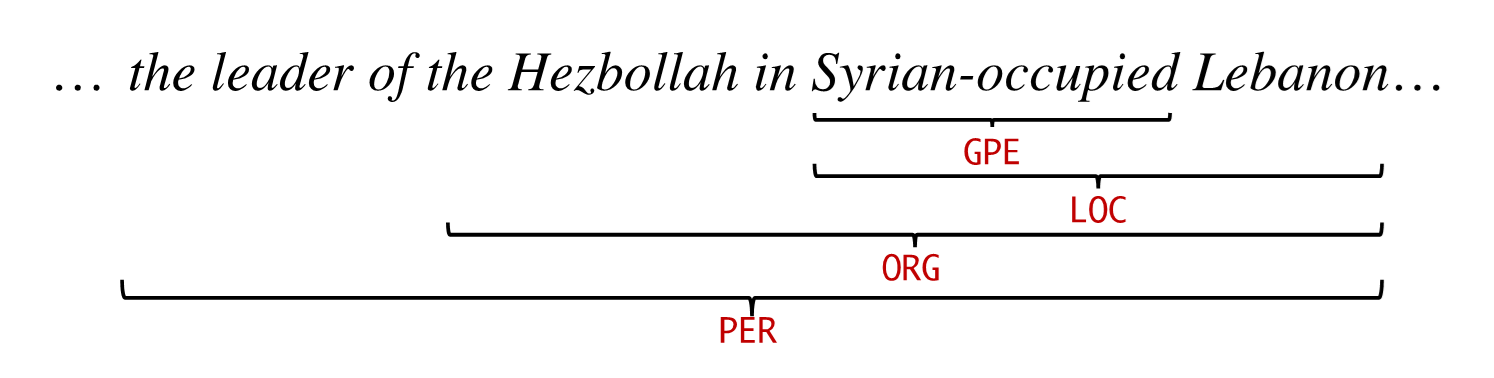}
    \caption{An example of nested mentions in ACE2004.}
    \label{fig:example}
    \end{center}
\end{figure}

Nonetheless, the standard method for classic NER treats the problem as a sequence labeling task which has difficulty recognizing entities with nested structures directly \cite{alex2007recognising, lu2015joint, katiyar2018nested}. With that in mind, various approaches to recognizing nested entities have been proposed.
From hyper-graph based methods \cite{lu2015joint,wang2018neural,marinho2019hierarchical} which design expressive tagging schemas, to span-based methods which classify the categories of sub-sequences \cite{sohrab2018deep, luan2019general, xia2019multi, fisher2019merge}.
In order to improve output quality, most more recent approaches to nested NER adopt structures that require the enumeration or heuristic traversal of all sub-sequences, which leads to inefficiency, and lack effective use of boundary information, which is very significant for nested entities.
Recently, \citet{zheng2019boundary} and \citet{tan2020boundary} explore using boundary knowledge to enhance recognition of nested entities.
But both focus only on entity start/end information, which face limitations in handling long entity spans, the interaction of entity start/end(s), and lack region information.
Moreover, decoupling the different nested levels of entity information remains a big problem in the nested NER task.
For example, the spans \textit{"the leader of the Hezbollah in Syrian-occupied Lebanon"}, \textit{"the Hezbollah in Syrian-occupied Lebanon"}, and \textit{"Syrian-occupied Lebanon"} all share the same end token (see Fig \ref{fig:example}).
Shared contextual representations tend to focus on the outermost entity type (in the above case, PER).

In this paper, we propose a novel joint entity mention detection and typing model via prior boundary knowledge, \textit{BoningKnife}, which can carve entity boundaries accurately and tease out type information more precisely.
Our model consists of two main components, MentionTagger and TypeClassifier, which are jointly trained with a common encoder representation and a shared dual-info attention layer.
MentionTagger performs mention detection by better leveraging boundary knowledge beyond just entity start/end to better handle nesting levels and longer spans. This improved representation of boundary knowledge both addresses limitations of previous systems and allows the generation of high quality mention candidates, which are critical for the overall system efficiency.
TypeClassifier then utilizes a new two-level attention mechanism to decouple different nested level representations and better distinguish entity types.
Moreover, the offshoots of MentionTagger entity token detection are further leveraged in the dual-info attention layer to improve joint training performance.


Experimental results on three datasets show that our approach achieves significant improvements over state-of-the-art methods across multiple nested NER datasets.
Further analysis and case studies demonstrate the effectiveness of each component in the model and its different sub-task strategies of mention detection and typing attention layers. Moreover, our approach also achieves higher efficiency without drop in quality. 



\section{Related Work}
\label{sec:Related Work}

Traditionally, most approaches formalize the NER task as a sequence labeling problem, which assigns a single label to each token in a sentence.
\cite{shen2003effective,zhang2004enhancing,zhou2006recognizing} adopt bottom-up methods, which performs entity recognition from inner to outer mentions, following hand-crafted rules. 


\citet{lu2015joint} introduces the idea of using a graph structure to connect tokens with multiple entities. While \cite{muis2018labeling,wang2018neural,katiyar2018nested,wang2019combining} propose hypergraphs and different methods to utilize graph information for nested NER. 

Transition-based models, which assemble a shift-reduce structure to detect a nested entity, have also been proposed. \citet{wang2018transition} builds a forest structure based on shift-reduce parsing.
\citet{marinho2019hierarchical} uses a stack structure to construct the transition-shift-reduce model. And 
\citet{ju2018neural} proposes a dynamically stacked multiple LSTM-CRF model to recognize the entity in an inside-out manner until no outer entity is plucked.

Span-based methods are another class of methods to recognize nested entities by classifying sub-sequences \cite{xia2019multi}. \citet{luan2019general} proposes a graph-based model which leverages entity linking to improve NER performance. \citet{fisher2019merge} introduces a merge-and-label method which uses nested entity hierarchy features.
\citet{strakov2019neural} views the nested NER task as a seq2seq generator problem, in which the input is a list of sentences and output target entities list.
\citet{shibuya2019nested} introduces an improved CRF model that recursively decoder the entity from outside to inside.

However, most previous methods need to traverse all sub-sequences and lack boundary knowledge, which is significant for nested entities. To try and mitigate such shortcomings, 
\citet{zheng2019boundary} defines a boundary detection task to generate a mention candidate set based on the entity start/end, followed by typing all mentions in the candidate set. And \citet{tan2020boundary} splits boundary information into two sub-tasks (entity start and entity end), before classifying candidates.
While both works use boundary knowledge, they focus only on entity start/end, which does not fully represent boundary information and lead to issues such as not handling long spans well. 
\textit{BoningKnife} jointly trains entity mention detection and typing modules and utilizes an extended representation of boundary knowledge to address such limitations.

\section{Problem and Methodology}
\label{sec:Model}

\begin{figure*}[htb!]
    \begin{center}
    \includegraphics[width=\textwidth]{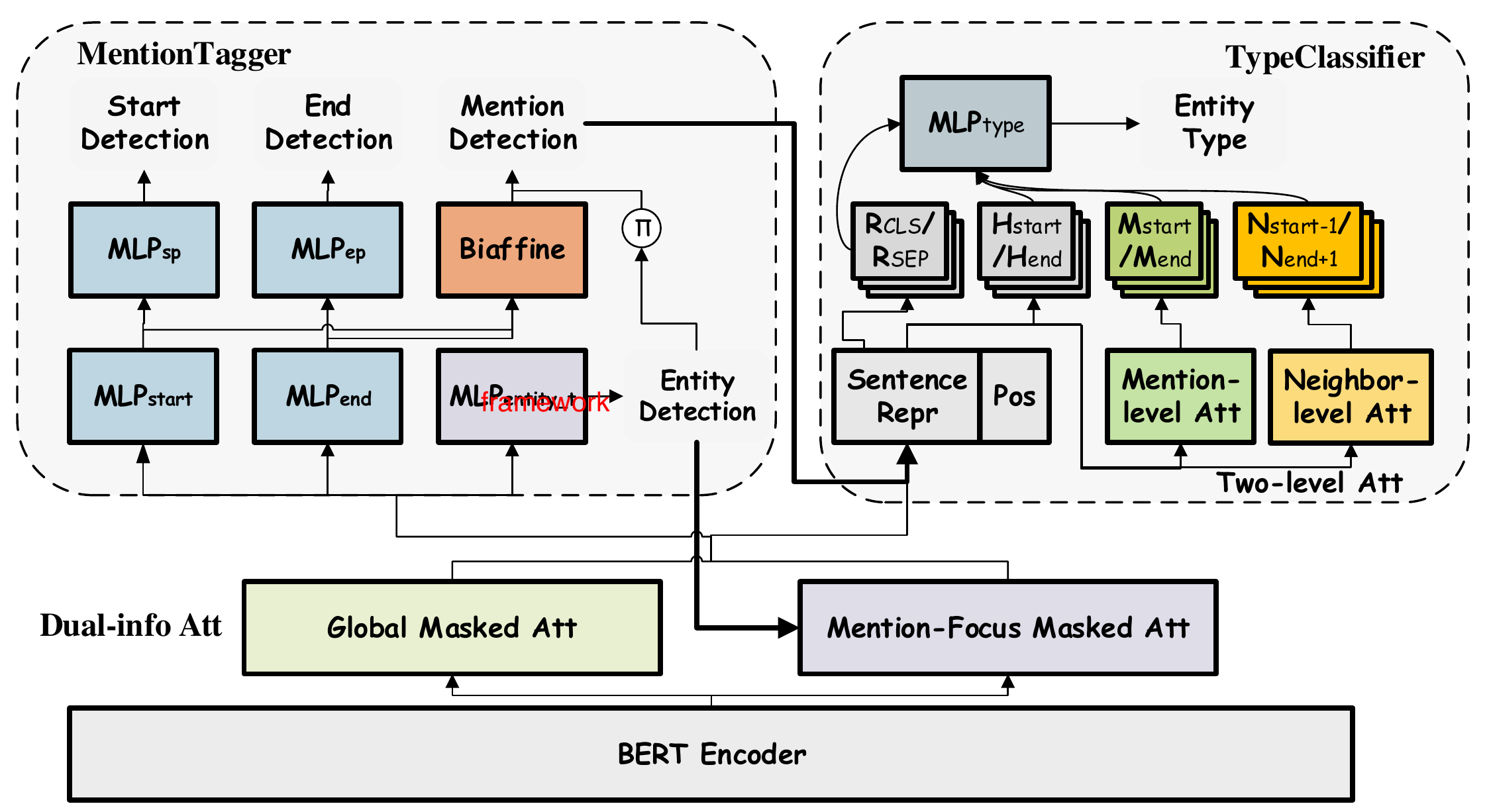}
    \caption{Framework of the proposed \textit{BoningKnife}, including Encoder Layer, MentionTagger, and TypeClassifier.}
    \label{fig:framework}
    \end{center}
\end{figure*}

In this section, we define the nested NER task, and then elaborate on our proposed solution.
Fig \ref{fig:framework} illustrates the framework of the proposed (\textit{BoningKnife}).
Specifically, we jointly train tagger and classifier, where the former (MentionTagger) extracts potential mention spans and generates mention candidates by leveraging an improved representation of boundary knowledge, and the later (TypeClassifier) classifies mention candidates into predefined entity types.

\subsection{Problem Statement}
\label{subsec:problem}

Let $\v s\in \mathbb{S}$ denote sentence data and $\v c \in \mathbb{C}$ denote entity label data, where $\mathbb{S}$ and $\mathbb{C}$ are the vector space of sentences and labels.
Given a sentence $\v s= [t_{i}]_{1 \le i \le N}$, where $N$ means the length of sentences $\v s$ and $t_i$ represents the $i$-th token of $\v s$, the NER task object is to extract all semantic elements $(l_e, r_e, c_e) \triangleq \v e \in \mathbb{E}$ where $l_e, r_e$ are the element start/end indices, $c_e$ means the element corresponding to a predefined label, and $\mathbb{E}$ is the entity space.

Essentially, nested NER aims to learn a $\mathcal{R}^{N \times N \times |\mathbb{C}|}$ space representation (only for the upper triangular matrix), where $|\mathbb{C}|$ is the size of entity categories and non-entity, and each value in the matrix represents the span type probability $\mathcal{P}(\v e|\v s)$.
We decompose the target probability $\mathcal{P}(\v e|\v s)$ into the product of two conditional probabilities (detection and typing) with a latent parameter (span).
\begin{equation}
\begin{aligned}
    \mathcal{P}(\v e|\v s) = \sum_i^{|\mathbb{E}|} {\mathcal{P}(l_i,r_i|\v s) \cdot\mathcal{P}(e|\v s,l_i,r_i)}
\end{aligned}
\end{equation}

In the formula, we discard the term with a small span probability $\mathcal{P}(l_i,r_i|\v s)$ in order to reduce the amount of calculations.

\subsection{Encoder Layer}
\label{subsec:encoder_layer}

\begin{figure}[t]
    \begin{center}
    \includegraphics[width=0.5\textwidth]{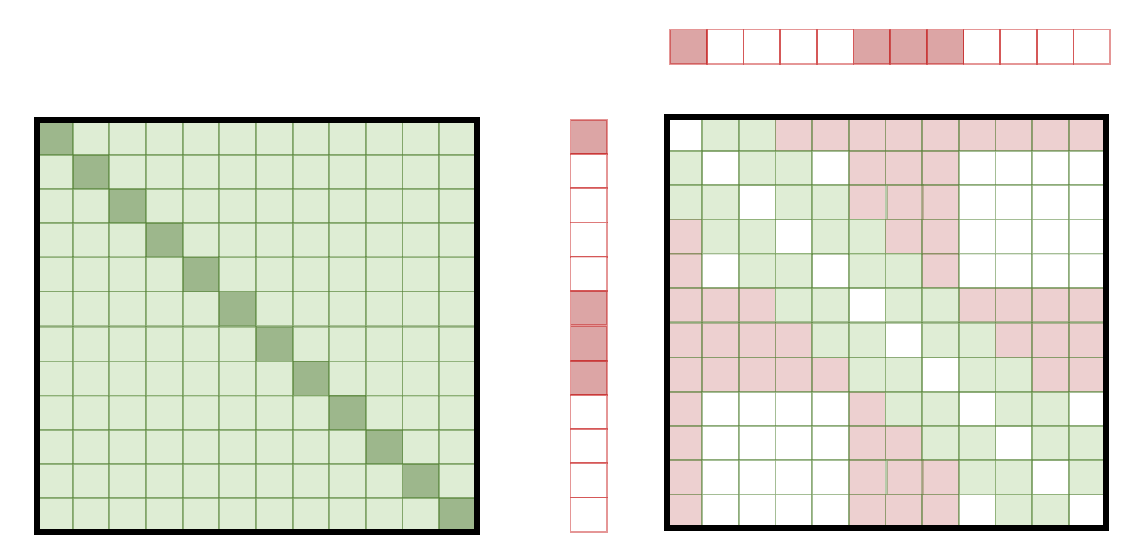}
    \caption{Mask matrix of Dual-info attention (windows size is 2). \textbf{Left}: global mask. \textbf{Right}: mention-focus mask which fuses "entity detection" subtasks and local context.}
    \label{fig:masked}
    \end{center}
\end{figure}

Because of the importance of context to entities, it is necessary to infuse information from different nested entities into one token. We propose a Dual-info attention structure to obtain entity semantic knowledge from both the token itself and others. This Dual-info attention representation is the input of two main sub-components of the model.

For the attention architecture, we use a pre-LayerNorm residual connection and multi-head attention mechanism.
\begin{equation}
    \attblok(\v s, \mask_s) = \multihead(\layernorm(\v s), \mask_s) + \v s
\label{eq:attln}
\end{equation}
where $\mask_s \in \mathbb{R}^{N\times N}$ is the attention mask matrix.


\textbf{Dual-info Attention} consists of {Global Masked attention} and {Mention-focus Masked attention} layers.
Fig \ref{fig:masked} shows the masked matrix example of Dual-info attention based on $\mathrm{BERT_{base}}$ \cite{devlin2018pretraining}.
We use $\v b_i$, the BERT representation of the $i$-th token in the sentence, to compute the Dual-info attention representation $\v R_i$.
\begin{equation}
    \v{R}_i= \linear([\attblok(\v b_i,\mask_{\t{global}}); \attblok(\v b_i,\mask_{\t{focus}})])
\end{equation}

{Global Masked Attention} considers every token from the same sentence, which makes the representation more contextual, 
while {Mention-focus Masked Attention} uses entity detection from MentionTagger (Sec \ref{subsec:MentionTagger}) and local context to construct attention weights.
For tokens not in mention candidates, we encode them from the representation of mention candidates’ tokens and local context.
Otherwise, we encode mention candidates’ tokens by calculating the attention weighted sum of all tokens except itself.
This approach tries to emphasise information related to: entity to entity, token to entity, and entity types.
The ablation experiments (Sec \ref{sec:ablation}) and attention discussion (Sec \ref{sec:attention_weight}) further showcase its effects.

\subsection{MentionTagger}
\label{subsec:MentionTagger}

The mention tagger module aims to extract entity mention candidates and compute their corresponding probabilities in a sentence. 

It is onerous to learn the mention detection matrix in $\mathbb{R}^{N\times N \times 2}$ space directly.
A basic idea is to traverse all sub-sequences based on a shared representation.
Building a high dimensional matrix concatenating the representation of entity's start/end token.
However, this method misses the interaction between start and end tokens, and increases the risk of over-fitting in the training stage.

The other extreme is to treat entity boundary information as two completely independent variables, like recent mention detection models \cite{zheng2019boundary,tan2020boundary} that only consider entity start/end tokens, which lack enough region contextual information. 
MentionTagger circumvents these two problems and utilizes three types of boundary information: entity start/end token, entity token itself, and mention region; which we term \textit{prior boundary knowledge}.
To infuse prior boundary distribution, we use three sub-tasks (start/end detection, entity detection, and mention detection) in training the tagger module.

\textbf{Start/End detection}
Inspired by the biaffine model \cite{dozat2016biaffine}, we use two MLPs ($\mlp_{\t{start}}$ and $\mlp_{\t{end}}$) to get a low dimension start/end representation $\v h_{\t{start}}, \v h_{\t{end}}$ and compute the span representation $\v S_e(i)$ for span $\v e(l_i, r_i)$.
We also use a start/end detection sub-task to enhance start/end representation $\v h_{\t{start}}, \v h_{\t{end}}$ and apply two other MLPs($\mlp_{\t{startPoint}}$ and $\mlp_{\t{endPoint}}$) to project $\v h_{\t{start}}$/$\v h_{\t{end}}$ to the category space.
For span $\v e(l_i, r_i)$, these are:
\begin{equation}
\begin{aligned}
    \v h_{\t{start}}(i)= & \mlp_{\t{start}}\left(\v R_{l_i}\right) \\
    \v h_{\t{end}}(i)= & \mlp_{\t{end}}\left(\v R_{r_i}\right) \\
\end{aligned}
\end{equation}
where $\v R_{l_i}$ and $\v R_{r_i}$ are the Dual-info Attention representation of the span $\v e(l_i, r_i)$'s start/end token.

And the span vector $\v S_e(i)$ generated from the low dimension start/end representation $\v h_{\t{start}}, \v h_{\t{end}}$,
\begin{equation}
    \v S_e(i)= \v h_{\t{start}}(i) \v U_{\t{span}} \v h_{\t{end}}(i) + \v h_{\t{start}}(i)\v u_{\t{start}} + \v h_{\t{end}}(i)\v u_{\t{end}} + \v b_{\t{span}} \\
\end{equation}
where $\v U_{\t{span}} \in \mathbb{R}^{N\times N \times d_{\t{span}}}$,
$u_{\t{start}} \in \mathbb{R}^{d_{\t{low}} \times d_{\t{span}}}$,
$u_{\t{end}} \in \mathbb{R}^{d_{\t{low}} \times d_{\t{span}}}$ are self-learned parameters.

\begin{equation}
    \begin{aligned}
        \mathcal{P}_{\theta}({\v Y}_{\t{start}}) &= \softmax(\mlp_{\t{startPoint}}(\v h_{\t{start}}(i))) \\
        \mathcal{P}_{\theta}({\v Y}_{\t{end}}) &= \softmax(\mlp_{\t{endPoint}}(\v h_{\t{end}}(i))) \\
    \end{aligned}
\end{equation}
where $\mathcal{P}_{\theta}({\v Y}_{\t{start}})$ and $\mathcal{P}_{\theta}({\v Y}_{\t{end}})$ are the output probability of the start/end detection sub-task.

\begin{equation}
\begin{aligned}
    \mathcal L_{\t{start}} &= \frac{1}{N} \sum_{i \in [1, N]} -\log{\mathcal{P}_{\theta}(Y_{\t{start},i} = y_{\t{start}, i})} \\
    \mathcal L_{\t{end}} &= \frac{1}{N} \sum_{i \in [1, N]} -\log{\mathcal{P}_{\theta}(Y_{\t{end},i} = y_{\t{end}, i})} \\
\end{aligned}
\end{equation}
where $y_{\t{start},i}$ are the ground truth labels of the start/end detection sub-task and $\mathcal L_{\t{start}}$, $\mathcal L_{\t{end}}$ are the training losses of the start/end sub-task, respectively.


\textbf{Entity detection}
Notice that only using the start/end information does not define a span boundary. Two high probability start $i$/end $j$ don't mean the probability of span $\v e(l_i, r_i)$ is high.
For example, in the sentence \textit{"Joe went to school."}, the probability of start token\textit{"Joe"} and end token \textit{"school"} are both high, but the probability of span \textit{"Joe went to school"} being an entity is low.
Applying the entity detection information (verdict token w/o belonging to at least one entity) can help address this problem. We reduce the span probability with large spacing by accumulating the entity detection probability values.

For the token index $k$ in a sentence, we have:
\begin{equation}
\begin{aligned}
    \mathcal{P}_{\theta}({\v Y}_{\t{enityDetection}}) = \softmax(\mlp_{\t{enityDetection}}(\v R_{k})) \\
\end{aligned}
\end{equation}
where $\mathcal{P}_{\theta}({\v Y}_{\t{enityDetection}})$ means the output probability of token $k$ belonging to a span.

In this sub-task, the entity detection loss function $\mathcal L_{\t{entityDetection}}$ is defined as:
\begin{equation}
    \mathcal L_{\t{entityDetection}} = \frac{1}{N} \sum_{i \in [1, N]} -\log{\mathcal{P}_{\theta}(Y_{\t{enityDetection},i} = y_{\t{enityDetection}, i})} \\
\end{equation}
where $y_{\t{entityDetection},i}$ are the ground truth labels of the entity detection sub-task.

\textbf{Mention detection}
Using the span representation $\v S_{e}(i)$ and entity detection probability $\mathcal{P}_{\theta}({\v Y}_{\t{enityDetection}})$, we compute the mention detection probability for all sub-sequences in the sentence.
For the span $\v e(l_i, r_i)$, these are:
\begin{equation}
\begin{aligned}
    \mathcal{P}_{\theta}({\v Y}_{\t{mention}}) = \softmax(\mlp_{\t{mention}}(\v S_{e}(i) \cdot \prod \limits_{k \in[l_i,r_i]} \mathcal{P}_{\theta}({\v Y}_{\t{enityDetection}}))) \\
\end{aligned}
\end{equation}
where $\mathcal{P}_{\theta}({\v Y}_{\t{mention}})$ means the output probability of the mention detection sub-task.

The mention detection loss function $\mathcal L_{\t{mention}}$ is calculated as follows:
\begin{equation}
\begin{aligned}
    \mathcal L_{\t{mention}} = \frac{1}{N^2} \sum_{i \in [1, N]} \sum_{j \in [1, N]} -\log{\mathcal{P}_{\theta}(Y_{\t{mention},i,j} = y_{\t{mention}, i,j})} \\
\end{aligned}
\end{equation}
where $y_{\t{entityDetection},i}$ are the ground truth labels of the mention detection sub-task.
After getting the mention probability $\mathcal{P}_{\theta}({\v Y}_{\t{mention}})$ of each mention pair $\v e(l_i,r_i)$, we use  a threshold hyper-parameter to generate the mention candidate $\mathbb{M}_{\t{candidate}}$.

MentionTagger not only outputs a mention candidate set $\mathbb{M}_{\t{candidate}}$ as the input of TypeClassifier, but the offshoots of its internal entity token detection $\mathcal{P}_{\theta}({\v Y}_{\t{enityDetection}})$ are fed to the shared Dual-info attention layer as a mention-focus mask matrix $\mask_{\t{focus}}$, improving its representation of entity semantic information.

\subsection{TypeClassifier}
\label{subsec:TypeClassifier}

After obtaining mention candidates from MentionTagger, TypeClassifier aims to predict the probability of entity types for each candidate.
We utilize a mention decoupling layer (MDL) to focus on the current mention semantic information, including two-level attention and a four-level representation (see Fig \ref{fig:framework}), for each span.

We apply a dimensional position embedding $\v {P}_i$ over the Dual-info Attention encoder embedding $\v{R}_i$, as position-wise token representation $\v{H}_i$, and consider $\v{H}_i$ as the input to the two-level attention component.
\begin{equation}
    \v{H}_i= \linear([\v R_i; \v P_i])
\end{equation}
where $\v P_i$ is a learnable position embedding. 

\textbf{Two-level attention}
combines mention-level attention $\mentionatt(\v s)$ (to make semantic information focus on internal spans)
and neighbor-level attention $\neighboratt(\v s)$ (which 
emphasizes the knowledge from span boundaries and contextual tokens). 
Both utilize have different mask matrices. Mention-level attention can only see the mention tokens, while neighbor-level attention can only see the remaining tokens.
This attention is defined as:
\begin{equation}
\begin{aligned}
    \mentionatt(\v s) &= \attblok(\v{H},\mask_{\t{mention}})\\
    \neighboratt(\v s) &= \attblok(\v H,\mask_{\t{neighbor}})\\
\end{aligned}
\end{equation}
where $\mask_{\t{mention}},\mask_{\t{neighbor}}$ are the difference mask matrices, and $\attblok$ is the attention architecture mentioned in eq (\ref{eq:attln}).

\textbf{Four-level representation}
For each span, we align and combine four fine-grained representations, thereby improving the model's understanding of entity boundaries and entity semantics.
Taking the span $\v e(l_i, r_i)$, the detail representation encompasses:

\begin{itemize}
\item Sentence-level Representation,
\begin{equation}
    \v E_{\t{sentence}}(\v s) = \linear([\v R_{\t{[CLS]}}; \v R_{\t{[SEP]}}])
\end{equation}
\item Position-wise Token Representation,
\begin{equation}
    \v E_{\t{position}}(\v s) = \linear([\v H_{\t{start}}; \v H_{\t{end}}])
\end{equation}
\item Mention-level Representation,
\begin{equation}
    \v E_{\t{mention}}(\v s) = \linear([\mentionatt_{\t{start}}; \mentionatt_{\t{end}}])
\end{equation}
\item Neighbor-level Representation,
\begin{equation}
    \v E_{\t{neighbor}}(\v s) = \linear([\neighboratt_{\t{previous}}; \neighboratt_{\t{next}}])
\end{equation}
\end{itemize}
where $X_{\t{[CLS]}},X_{\t{[SEP]}},X_{\t{start}},X_{\t{end}},X_{\t{previous}}, and X_{\t{next}}$ are the [CLS], [SEP], start, end, previous mention, and next mention token representations, respectively.

These four different features are combined into TypeClassifer's representation:
\begin{equation}
    \v T(\v s) = \linear([\v E_{\t{sentence}};\v E_{\t{position}};\v E_{\t{mention}};\v E_{\t{neighbor}}])
\end{equation}

In the same way, the predicted classification probability is output through the Softmax function.
\begin{equation}
\begin{aligned}
    \mathcal{P}_{\theta}({\v Y}_{\t{type}}) = \softmax(\mlp_{\t{type}}(\v T(\v s)))
\end{aligned}
\end{equation}

The corresponding loss function $\mathcal L_{\t{type}}$ is:
\begin{equation}
\begin{aligned}
    \mathcal L_{\t{type}} = \frac{1}{|\mathbb{M}_{\t{candidate}}|} \sum_{i \in [1, |\mathbb{M}_{\t{candidate}}|]} -\log{\mathcal{P}_{\theta}(Y_{\t{type},i} = y_{\t{type}, i})} \\
\end{aligned}
\end{equation}

\subsection{Optimization Objective}
\label{subsec:optimizationobjective}

In MentionTagger, we joint optimize the above detection sub-tasks. 
All losses are based on cross entropy loss. To balance difference loss, we apply a focal loss style self-adjusting weight strategy.
For larger losses, the weights will be correspondingly scaled up to improve the learning process.
\begin{equation}
    \mathcal{L}_{\t{mentionTagger}} = \alpha_{\t{start}}\mathcal L_{\t{start}} + \alpha_{\t{end}}\mathcal L_{\t{end}} + \alpha_{\t{entityDetection}}\mathcal L_{\t{entityDetection}} + \alpha_{\t{mention}}\mathcal L_{\t{mention}}  \\
\end{equation}
where, 
\begin{equation}
\begin{aligned}
    \learnscore_{i} &= (1-\exp(-\mathcal{L}_i)) ^ \beta \\\\
    \alpha_{i} &= \frac{\learnscore_{i}}{\sum_{j}{\learnscore_{j}} } \\
\end{aligned}
\end{equation}
where $\learnscore_{i}$ is a score to judge degree of sub-task training, which borrows from focal loss~\cite{lin2017focalloss}
and $\alpha_{i}$ is a normalization version of $\learnscore_{i}$,  $j\in\{\t{start},\t{end},\t{entityDetection},\t{mention}\}$.

We jointly train MentionTagger and TypeClassifier as a multi-task process alternately, where the shared representation layer and the entity detection prediction results in Dual-info attention enhance the connections between the two components.
The overall optimization goal is:
\begin{equation}
    \mathcal{L}_{\t{train}} = \mathcal L_{\t{mentionTagger}} + \mathcal L_{\t{type}}  \\
\end{equation}

\section{Experiments}
\label{sec:Experiments}

\subsection{Datasets}
\label{sec:datasets}

We evaluate our model on three nested NER datasets, ACE2004 \cite{doddington2004automatic}, ACE2005 \cite{walker2006ace}, and NNE \cite{dblp2019nne}; using the same splits as previous work \cite{lu2015joint,wang2018neural,lin2019sequence,dblp2019nne}.\footnote{ACE2004 / ACE2005 as in \url{https://statnlp-research.github.io/publications/}, and NNE as in \url{https://github.com/nickyringland/nested_named_entities}}.
Table \ref{tab:stats_nested} shows the proportions of nested entities in the datasets range from 30.80\% to 66.14\%. ACE2004 and ACE2005 include 7 entity types, while NNE has 114 entity types.
We report precision, recall, and micro-F1 metrics for all experiments.

\begin{table}[htbp]
\centering
\vskip 0.1in
\begin{tabular}{@{}lccc@{}}
\toprule
& ACE2004 & ACE2005 & NNE \\ \midrule
Documents & 443 & 464 & 2,312 \\
Sentences & 8,507 & 9,311 & 49,208 \\
Mentions  & 27,753 & 31,102 & 279,795 \\
Entity overlaps & 9,767 & 9,579 & 185,054 \\
Overlap ratio & 35.19\% & 30.80\% & 66.14\% \\
\% of overlaps over all sub-sequences & 1.11\% & 1.30\% & 1.47\% \\
\bottomrule
\end{tabular}%
\caption{Statistics of the ACE2004, ACE2005, and NNE nested NER datasets.}
\label{tab:stats_nested}
\end{table}

\subsection{Baseline and Experimental Settings}
\label{sec:baseline}

We compare \textit{BoningKnife} with a set of representative models and the recent state of the art.

\begin{itemize}
\item \citet{wang2018neural}, a graph-based model using LSTM to learn a feature encoder;
\item \citet{xia2019multi}, which is a detect-classify model without boundary knowledge;
\item \citet{luan2019general}, a graph-based model which leverages entity linking to improve NER;
\item \citet{fisher2019merge}, a merge-and-label model with hierarchical features;
\item \citet{strakov2019neural}, which treats the nested NER task as a seq2seq problem;
\item \citet{shibuya2019nested}, which extracts nested entities recursively with CRF;
\item \citet{tan2020boundary}, which combines entity start/ end probabilities.
\end{itemize}

Moreover, we utilize $\mathrm{BERT_{base}}$ itself as a lower-bound baseline for the BERT-based models.\footnote{NNE results for \cite{shibuya2019nested} are reported by using their public code in \url{https://github.com/yahshibu/nested-ner-2019-bert}.}

We perform a random search strategy for hyperparameter optimization and select the best settings on the development sets.
We initialize the loss weight parameter as $\beta = 0.5$ in MentionTagger and set the max neighbor window size to 128 in neighbor-level attention block $\neighboratt$ to control memory size.
The hidden sizes of low dimension start/end representation $\v h_{\t{start}},\v h_{\t{end}}$ are 84, the number of attention heads is 16, and the windows size of mention-focus attention mask matrix $\mask_{\t{focus}}$ is 2.
Except for $\mathrm{BERT_{base}}$, our model has around 24M parameters.
We employ AdamW \cite{loshilov2017adamw} as optimizer during training.
Experiments are repeated 5 times for different random seeds on each corpus.

\subsection{Results and Discussion}

\begin{table*}[!t]
   \centering
   \vskip 0.1in
   \begin{tabular}{@{}l|ccc|ccc@{}}
   \toprule
   \multirow{2}{*}{\textbf{Model}} & \multicolumn{3}{c|}{\textbf{ACE2004}} & \multicolumn{3}{c}{\textbf{ACE2005}} \\ \cmidrule(l){2-7} 
    & \textbf{P}(\%) & \textbf{R}(\%) & \textbf{F1}(\%) & \textbf{P} (\%) & \textbf{R}(\%) & \textbf{F1} (\%) \\ \midrule
      \citet{wang2018neural}      &78.0  & 72.4  & 75.1  &76.8  & 72.3  & 74.5  \\
      \citet{xia2019multi} [ELMO] &81.7  & 77.4  & 79.5  &79.0  & 77.3  & 78.2  \\
      \citet{luan2019general}     &   -  &   -   & 84.7  &   -  &   -   & 82.9 \\
      \hline
      {BERT} merge outmost \& inmost$^{\dag}$&80.55 & 79.23 & 79.88 &78.12 & 82.71 & 80.35 \\
      \citet{fisher2019merge}$^{\dag}$     &   -  &   -   &   -   &82.7  & 82.1  & 82.4  \\
      \citet{strakov2019neural}$^{\dag}$   &   -  &   -   & 84.4  &   -  &   -   & 84.3  \\
      \citet{shibuya2019nested}$^{\dag}$   &85.23 & 84.72 & 84.97 &83.30 $\pm$ 0.22 & 84.69 $\pm$ 0.37 & 83.99 $\pm$ 0.27 \\
      \citet{tan2020boundary}$^{\dag}$    &85.8 & 84.8 & 85.3 &83.8 & 83.9 & 83.9 \\
      \hline
      \textbf{BoningKnife}$^{\dag}$         &\textbf{85.98} $\pm$ 0.36  & \textbf{86.86} $\pm$ 0.39 & \textbf{86.41} $\pm$ 0.24 &\textbf{84.77} $\pm$ 0.31 & \textbf{86.16} $\pm$ 0.43 & \textbf{85.46} $\pm$ 0.32 \\
    \bottomrule
   \end{tabular}%
   \caption{Results of the proposed \textit{BoningKnife} and prior state-of-the-art methods over the ACE2004/2005 test sets. $^{\dag}$ denotes models utilizing $\mathrm{BERT}$. '-' denotes results not reported.}
   \label{tab:main_results}
\end{table*}
   
\begin{table}[!t]
\centering
\vskip 0.1in
\begin{tabular}{@{}lccc@{}}
\toprule
\textbf{Model} & \textbf{P}(\%) & \textbf{R}(\%) & \textbf{F1}(\%) \\ \midrule
 \citet{wang2018transition}   &77.4  & 70.1  & 73.6  \\
 $\mathrm{BERT}_{base}$ merge outmost \& inmost&80.43 & 74.94 & 77.59 \\
 \citet{wang2018neural}      &91.8  & 91.1  & 91.4  \\
 \citet{shibuya2019nested}\footnotemark &93.03  & 93.34  & 93.19  \\
 \hline
 \textbf{BoningKnife}        &\textbf{93.74} & \textbf{94.75} & \textbf{94.24}  \\
                             &($\pm$ 0.33) & ($\pm$ 0.24) & ($\pm$ 0.05) \\
\bottomrule
\end{tabular}%
\caption{Results of the proposed model and baselines over the NNE dataset.}
\label{tab:nne_result}
\end{table}


Tables \ref{tab:main_results} and \ref{tab:nne_result} report the results of our model and the different baselines on the ACE2004/ACE2005 and NNE datasets.
It can be seen that our proposed method outperforms all previous state-of-the-art methods, reaching 86.41, 85.46, and 94.24 in average micro-F1 score, on ACE2004, ACE2005, and NNE respectively.

Compared with the latest boundary-enhanced method \citep{tan2020boundary}, our method achieves 1.11 and 1.56 absolute point gains on ACE2004 and ACE2005 \footnote{Unfortunately \citet{tan2020boundary} does not report results over NNE and did not release their code for further experiments}.
The boost comes mainly from recall improvements (2.06 to 2.26 points). MentionTagger is able to produce more precise mention candidates, which allows TypeClassifier to focus on distinguishing entity types, instead of filtering candidates as not viable. The improved precision of MentionTagger is further evidenced in Table \ref{tab:time_compare}.


In the NNE corpus, \textit{BoningKnife} achieves 94.24 F1-score; an improvement of 1.05 points over the previous SOTA. We hypothesize that as NNE datasets has deeper nesting levels, \citep{tan2020boundary}'s approach leads to error transmission in their recursive encoding process.

\section{Analysis}
\label{sec:analysis}

\subsection{Ablation \& Flat/Nested Performance}
\label{sec:ablation}

\begin{table}[htbp]
\centering
\vskip 0.1in
\begin{tabular}{@{}lccc@{}}
\toprule
\textbf{Method} & \textbf{P}(\%) & \textbf{R}(\%) & \textbf{F1}(\%) \\ \midrule
\textbf{BoningKnife}          &$85.98$ & $86.86$ & $86.41$ \\
- w/o EntityDetection subtask  &$84.74$ & $85.95$ & $85.35$ \bf{(-1.06)} \\
- w/o Start/End sub-task &$85.71$ & $85.18$ & $85.44$ \bf{(-0.97)} \\
- w/o Neighbor-level attention   &$85.44$ & $86.56$ & $86.00$ \bf{(-0.41)} \\
- w/o Two-level attention   &$84.75$ & $86.43$ & $85.58$ \bf{(-0.83)} \\
- w/o Mention-focus attention   &$85.64$ & $86.63$ & $86.13$ \bf{(-0.28)} \\
- w/o MentionTagger Stage\footnotemark &$87.25$ & $83.86$ & $85.52$ \bf{(-0.89)} \\
\bottomrule
\end{tabular}%
\caption{Ablation study of the proposed \textit{BoningKnife} over the ACE 2004 test set, where numbers in parenthesis denote performance change.}
\label{table:ablation}
\end{table}

To validate the contributions and effectiveness of different components in
the proposed model, we introduce the following model variants to perform an ablation study:


Table \ref{table:ablation} highlights the performance contributions of each
component in our proposed model, and removing any of them will generally lead to substantial performance drops.
It can be seen that quality decreases significantly when either removing MentionTagger or its sub-tasks (entity token and start/end detection) sub-task, which indicates the proposed model makes effective usage of boundary knowledge (for example, to better handle long length entity spans).
Without the proposed two-level attention in TypeClassifier, it becomes 
harder for the model to separate nested information and assign the proper type for nested entities; even more so than removing only the neighbor-level component of the two-level attention . This further demonstrates the benefits of the two-level structure and its ability to combine clear boundary and local context information.
Lastly, while the effects of removing mention-focus mask attention are less prominent, it's still noticeable and removing this component leads to slower overall model convergence.
Furthermore, Table \ref{tab:span_result} reports the Flat/Nested performance across datasets. It can be seen that \textit{BoningKnife} excels in nested entities while remaining competitive on flat results; which further evidences the overall effectiveness of the model in leveraging boundary knowledge.

\footnotetext{Replacing MentionTagger with using entity start/end to generate mention candidates, similarly to \citet{tan2020boundary}.}




\begin{table}[htbp]
\centering
\vskip 0.1in
\begin{tabular}{@{}l|cc|cc|cc@{}}
\toprule
\multirow{2}{*}{\textbf{Method}} & \multicolumn{2}{c|}{\textbf{ACE2004}} & \multicolumn{2}{c|}{\textbf{ACE2005}} & \multicolumn{2}{c}{\textbf{NNE}} \\ \cmidrule(l){2-7} 
 & \textbf{Flat} & \textbf{Nested} & \textbf{Flat} & \textbf{Nested} & \textbf{Flat} & \textbf{Nested} \\ \midrule
\textbf{BoningKnife} & 84.32 & \textbf{87.10} & 84.54 & \textbf{86.23} & \textbf{84.45} & \textbf{94.73}   \\
-w/o MentionTagger & 83.56 & 86.27 & 84.16 & 85.45 & 83.89 & 93.65 \\
\citet{shibuya2019nested} & \textbf{84.45} & 85.14 & \textbf{84.86} & 83.13 & 84.26 & 93.67  \\
\bottomrule
\end{tabular}%
\caption{\textbf{Flat/Nested} F1-scores over the ACE2004, ACE2005, and NNE test sets. 
}
\label{tab:span_result}
\end{table}

\subsection{Time Complexity}
\label{sec:time_complexity}

\begin{table}[htbp]
\centering
\vskip 0.1in
\begin{tabular}{@{}l|cc|cc|cc@{}}
\toprule
\multirow{2}{*}{\textbf{Method}} & \multicolumn{2}{c|}{\textbf{ACE2004}} & \multicolumn{2}{c|}{\textbf{ACE2005}} & \multicolumn{2}{c}{\textbf{NNE}} \\ \cmidrule(l){2-7} 
 & \textbf{P} & \textbf{Time (s)} & \textbf{P} & \textbf{Time (s)} & \textbf{P} & \textbf{Time (s)} \\ \midrule
\textbf{BoningKnife} & 89.46 & 479 & 87.22 & 649 & 95.48 & 7841   \\
-w/o MentionTagger   & 31.54 & 839 & 32.58 & 1190 & 30.71 & 32758 \\
\midrule
\textbf{TimeRatio} &  & 1.75X &  & 1.83X & & 4.18X   \\
\bottomrule
\end{tabular}%
\caption{Comparison of mention precision and training time cost per epoch, between E2E system using MentionTagger and mention strategy from \citet{tan2020boundary}. }
\label{tab:time_compare}
\end{table}

Similarly to \citet{tan2020boundary}, our method substantially improves the time complexity over typical span-based methods by generating high-quality candidates, which greatly reduce complexity and training time. 
Span-based models, which require traversing all sub-sequences, have 
$O(m \cdot n^2$) time complexity, where $m$ is the count of tags. 
Efficiently reducing the number of candidates is key in a two-step system like \textit{BoningKnife}, as span classifying time complexity is determined by the number of candidates in its input. 
To measure the speedup from our approach due to its improved candidate generation and provide a comparison with \citet{tan2020boundary}, we run two experiments: i) the complete \textit{BoningKnife} system and ii) \textit{BoningKnife} - MentionTagger + mention strategy in \citet{tan2020boundary}. The experiments were run on a Ubuntu 16.04.6 server with Intel Xeon CPU E5-2690v3 @ 2.60GHz
and one P100 GPU.


Table \ref{tab:time_compare} reports the comparison between both experiments. We can see that our method provides significant speedup over the simpler modeling of boundary knowledge approach, especially with deeper nesting levels. \textit{BoningKnife} is 1.75x, 1.83x, and 4.18x faster in ACE2004, ACE2005, and NNE, respectively, while still achieving higher quality.


\subsection{Case Study and Attention Weight Visualization}
\label{sec:case_study}

\begin{table*}[t!]
\vspace{.1in}
\footnotesize \centering
\begin{tabular}{@{}rll@{\;\;}l@{\;\;}lllll@{}} \toprule
\multicolumn{9}{l}{\shortstack[l]{\textbf{Sentence:} The \textbf{Coventry University researchers who report the findings in the British journal of sports medicine} say \\anxiety and depression are common among \textbf{those so injured}, possibly as a result of pain and impaired mobility.}} \\
\midrule
& \textbf{SPAN} & \textbf{$p_{start}$} & \textbf{$p_{end}$} & \textbf{$p_{mention}$} & \textbf{$p_{s}$} & \textbf{$p_{type}$} & \textbf{$y_{type}$} & \textbf{$y_{golden}$} \\
\midrule
(a) & The Coventry University researchers who report the & 1.000 & 0.999 & 1.000 & 0.942 & 1.000 & PER & PER \\
& findings in the British journal of sports medicine & & & &  & & & \\
(b) & Coventry University & 0.595 & 1.000 & 0.945 & 0.502 & 0.998 & ORG & ORG \\
(c) & who & 1.000 & 0.999 & 1.000 & 0.997 & 1.000 & PER & PER \\
(d) & those so injured & 1.000 & 0.483 & 0.953 & 0.045 & 1.000 & PER & PER \\
(e) & those so injured, possibly as a result of pain and & 1.000 & 0.000 & 0.000 & 0.573 & 0.000 & Non-entity & Non-entity \\
& impaired mobility & & & &  & & & \\
(f) & . & 0.000 & 0.000 & 0.000 & 0.501 & 0.000 & Non-entity & Non-entity \\
\bottomrule
\end{tabular}
\caption{An example where \textit{BoningKnife} leverages prior boundary knowledge to better predict nested entity type.
$p_s$ from the ablation experiment "w/o ED subtask".
}
\label{tab:predict_examples}
\end{table*} 

Table \ref{tab:predict_examples} shows an example of \textit{BoningKnife} prediction in ACE 2004.
Span (d), \textit{"those so injured"}, is a correct mention, but the probability of end token \textit{"injured"} is small. For S/E based methods like \cite{tan2020boundary,zheng2019boundary}, this span would likely be discarded, but in our method it is correctly identified.
Compared with $p_s$ only, the entity token detection knowledge reduces the number of high probability mentions (like (e) and (f)) inconsistent with the prior information, while not discarding very long entities, like mention (a).

\label{sec:attention_weight}

\begin{figure}[t]
    \centering
    \subfigure[Global Mask]{
        \begin{minipage}[t]{0.55\linewidth}
        \includegraphics[width=0.9\textwidth]{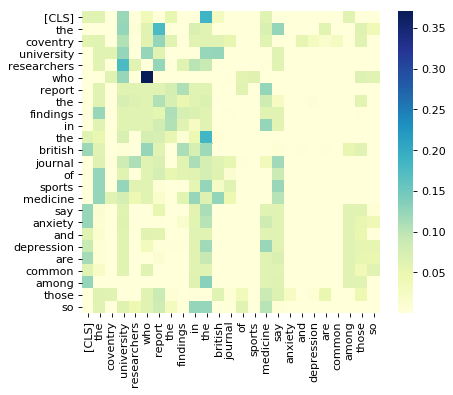}
        \end{minipage}%
    }%
    \subfigure[Mention-focus Mask]{
        \begin{minipage}[t]{0.55\linewidth}
        \includegraphics[width=0.9\textwidth]{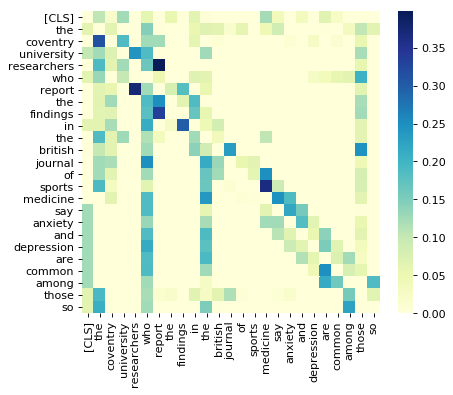}
        \end{minipage}%
    }%
    \centering
    \caption{Visualization of the Dual-info Attention weights for the case study sentence (Table \ref{tab:predict_examples}).}
    \label{fig:atten_weight}
\end{figure}

Fig \ref{fig:atten_weight} shows the Dual-info attention weights for the sentence in Table \ref{tab:predict_examples}.
The global attention weights all focus on common keywords like \textit{"university"}, \textit{"the"}. While the mention-focus attention, focus on specific token neighbors, like \textit{"report"} focusing on \textit{"researchers"} and \textit{"findings"}, which improve their semantic information. Also, additional tokens focus on the relevant entity tokens, like \textit{"who"} focusing on the same entity type word \textit{"those"} instead of on itself in global attention.

\section{Conclusion}
\label{sec:Conclusion}

In this paper we propose a novel joint entity mention detection and typing model via prior boundary knowledge for the nested NER task. 
The proposed method effectively incorporates prior boundary knowledge information to generate high quality mention candidates, which greatly improves efficiency of the whole system. By introducing a Dual-info attention layer at the mention classification stage, it facilitates mention decoupling and more accurate mention classification at different levels.
Experiments show that our system, \textit{BoningKnife}, achieves state-of-the-art results on three standard benchmark datasets; and an ablation study further demonstrates the effectiveness of its components.

\bibliographystyle{unsrtnat}
\bibliography{nestedNER}

\end{document}